\documentclass[journal]{IEEEtran}
\usepackage{cite}

\ifCLASSINFOpdf
  \usepackage[pdftex]{graphicx}
  \graphicspath{{Graphs/}}
  \DeclareGraphicsExtensions{.pdf,.jpeg,.png}
\else
\fi
\usepackage{glossaries} 

\newacronym{api}{API}{application programming interface}
\newacronym{dsp}{DSP}{digital signal processing}
\newacronym{dvfs}{DVFS}{dynamic voltage frequency scaling}
\newacronym{hal}{HAL}{hardware abstraction layer}
\newacronym{mac}{MAC}{multiply-accumulate}
\newacronym{ml}{ML}{machine learning}
\newacronym{muapt}{MUAPT}{motor unit action potential train}
\newacronym{sdk}{SDK}{software development kit}
\newacronym{lp}{LP}{low-power}
\newacronym{nda}{NDA}{non-disclosure agreement}
\newacronym{os}{OS}{operating system}
\newacronym{fxp}{FxP}{fixed-point}
\newacronym{fp}{FP}{floating-point}
\newacronym{soa}{SoA}{state-of-the-art}
\newacronym{isa}{ISA}{Instruction Set Architecture}

\newacronym{eembc}{EEMBC}{Embedded Microprocessor Benchmark Consortium}
\newacronym{iot}{IoT}{Internet-of-Things}
\newacronym{iomt}{IoMT}{Internet-of-Medical-Things}
\newacronym{wcet}{WCET}{worst-case execution time}

\newacronym{seizdetcnn}{SeizureDetCNN}{seizure detector convolutional neural network}
\newacronym{seizdetsvm}{SeizureDetSVM}{seizure detector support vector machine}
\newacronym{coughdet}{CoughDet}{cough detector}
\newacronym{cwm}{CognWorkMon}{cognitive workload monitor}
\newacronym{gcl}{GestureClass}{gesture classifier}
\newacronym{hcl}{HeartBeatClass}{heartbeat classifier}
\newacronym{ecl}{EmotionClass}{emotion classifier}
\newacronym{bpfree}{Bio-BPfree}{biological back-propagation-free}
\newacronym{semg}{sEMG}{surface electromyography}
\newacronym{ecg}{ECG}{electrocardiogram}
\newacronym{eeg}{EEG}{electroencephalogram}
\newacronym{gsr}{GSR}{Galvanic skin response}
\newacronym{ppg}{PPG}{Photoplethysmography}
\newacronym{st}{ST}{skin temperature}

\newacronym{bss}{BSS}{blind source separation}
\newacronym{cnn}{CNN}{convolutional neural network}
\newacronym{rp}{RP}{random projections}
\newacronym{rf}{RF}{random forest}
\newacronym{mf}{MF}{morphological filter}
\newacronym{ma}{MAVG}{moving average}
\newacronym{bpf}{BPF}{band-pass filter}
\newacronym{lpf}{LPF}{low-pass filter}
\newacronym{rms}{RMS}{root-mean-square}
\newacronym{fft}{FFT}{fast Fourier transform}
\newacronym{relen}{Rel-En}{relative energy}
\newacronym{dnn}{DNN}{deep neural network}
\newacronym{fcn}{FCN}{fully-convolutional network}
\newacronym{svm}{SVM}{support vector machine}
\newacronym{ica}{ICA}{independent component analysis}
\newacronym{mlp}{MLP}{multilayer Perceptron}
\newacronym{knn}{KNN}{k-nearest neighbors}
\newacronym{mfcc}{MFCC}{mel-frequency cepstral coefficients}
\newacronym{ecgfpdel}{ECG-FPDEL}{ECG fiducial points delineation}
\newacronym{avg}{AVG}{average}
\newacronym{reward}{REWARD}{relative-energy-based wearable R-peak detection}
\newacronym{rri}{RRI}{R-peak interval}
\newacronym{edr}{EDR}{ECG-derived respiration}
\newacronym{hrv}{HRV}{heart-rate variability}
\newacronym{lpc}{LPC}{linear predictive coefficients}
\newacronym{plomb}{PLOMB}{Lomb-Scargle periodogram}
\newacronym{adam}{ADAM}{Adam optimizer}
\newacronym{convgrad}{CONV-GRAD}{convolutional block gradients}
\newacronym{dct}{DCT}{discrete cosine transform}
\newacronym{pt}{PT}{Pan-Tompkin}
\newacronym{blr}{BLR}{blink removal}
\newacronym{iir}{IIR}{infinite impulse response}
\newacronym{ffe}{FFE}{frequency feature extraction}

\newacronym{dma}{DMA}{direct memory access}
\newacronym{fc}{FC}{fabric controller}
\newacronym{fpu}{FPU}{floating-point unit}
\newacronym{mcu}{MCU}{micro-controller unit}
\newacronym{pmu}{PMU}{power management unit}
\newacronym{spi}{SPI}{serial peripheral interface}
\newacronym{adc}{ADC}{analog-to-digital converter}
\newacronym{imu}{IMU}{inertial measurement unit}
\newacronym{ic}{IC}{integrated circuit}
\newacronym{ldo}{LDO}{low-dropout regulator}

\usepackage{siunitx} 
\usepackage{soul}    
\usepackage{booktabs}
\usepackage{float}
\usepackage{subcaption}
\usepackage{multirow} 
\usepackage{makecell}

\usepackage{xcolor}
\definecolor{myColor}{RGB}{0, 0, 0}
\usepackage{soul}
\begin{document}
%
\title{BiomedBench: A benchmark suite of TinyML biomedical applications for low-power wearables}
%
%
%

\author{Dimitrios~Samakovlis$^1$, Stefano~Albini$^1$, Rubén~Rodríguez~Álvarez$^1$, Denisa-Andreea~Constantinescu$^1$, Pasquale~Davide~Schiavone$^1$, Miguel~Peón-Quirós$^2$, David~Atienza$^1,^2$\\$^1\,$Embedded Systems Laboratory, École Polytechnique Fédérale de Lausanne (EPFL), Switzerland\\$^2\,$EcoCloud Center, École Polytechnique Fédérale de Lausanne (EPFL), Switzerland}
\maketitle

\begin{abstract}
The design of low-power wearables for the biomedical domain has received a lot of attention in recent decades, as technological advances in chip manufacturing have allowed real-time monitoring of patients using low-complexity \acrshort{ml} within the \si{\milli\watt} range. Despite advances in application and hardware design research, the domain lacks a systematic approach to hardware evaluation. In this work, we propose BiomedBench, a new benchmark suite composed of complete end-to-end TinyML biomedical applications for real-time monitoring of patients using wearable devices. Each application presents different requirements during typical signal acquisition and processing phases, including varying computational workloads and relations between active and idle times. Furthermore, our evaluation of five state-of-the-art low-power platforms in terms of energy efficiency shows that modern platforms cannot effectively target all types of biomedical applications. BiomedBench {\color{myColor}is} released as an open-source suite to {\color{myColor} standardize hardware evaluation and guide hardware and application design in the TinyML wearable domain.} 
\end{abstract}


%
\IEEEpeerreviewmaketitle

\section{Introduction}
%
%
%
%
\IEEEPARstart{W}{earable} devices promise to improve preventive medicine through continuous health monitoring of chronic diseases. To this end, we face the challenge of increasing their computational capability and energy efficiency to implement advanced biomedical algorithms on the edge, increase patient privacy, and reduce response latency while enabling seamless operation with small batteries and sparse recharging cycles. {\color{myColor}This paper explicitly focuses on the TinyML wearable domain where battery-powered devices operate in the \SI{}{\milli\watt} range to meet tight energy budgets while employing lightweight \gls{ml} models.} 

{\color{myColor} To increase research efficiency in the TinyML wearable domain, it is vital to facilitate a seamless synergy between software and hardware development efforts. First,} 
architectural design must be aligned with the characteristics of \gls{soa} applications to meet real-time and energy constraints. {\color{myColor}Second,} application developers must be aware of {\color{myColor}\gls{soa} algorithms and platforms} in the domain to minimize software development and deployment time. However, {\color{myColor} in the TinyML wearable domain}, we are missing in the \gls{soa} a set of representative end-to-end applications to guide the co-design process by standardizing hardware evaluation and unveiling the critical hardware and software design points. 

In response to these needs, we propose BiomedBench, a biomedical benchmark suite composed of end-to-end {\color{myColor}TinyML} applications aimed at low-power wearable devices. These applications feature diverse requirements during processing, idle, and signal acquisition that effectively represent the challenges in the domain. Furthermore, we demonstrate how to utilize BiomedBench to systematically evaluate and compare \gls{soa} platforms. To our knowledge, this is the first benchmark suite explicitly targeting the low-power {\color{myColor}wearable TinyML} domain, offering a systematic approach to software and hardware co-design. The contribution of BiomedBench is twofold:
\begin{itemize}
  \item {\color{myColor}It standardizes hardware evaluation in the TinyML wearable domain, offering a set of complete end-to-end biomedical applications, including the idle, acquisition, and processing phases. The variety of requirements present in the applications is representative of the multi-dimensional challenges in the domain.}

  \item {\color{myColor}It provides guidelines for future hardware and application design in the TinyML wearable domain.} Utilizing BiomedBench to compare \gls{soa} platforms unveils the critical design features that impact {\color{myColor}performance} for hardware designers and hints at the deployment platform selection for application designers. {\color{myColor}Overall, open-sourcing \gls{soa} applications accelerates future application development efforts.}
  
\end{itemize}

We organize this work as follows. In Section~\ref{sec:RelatedWork}, we {\color{myColor}identify the need for BiomedBench by comparing it} with existing biomedical benchmark suites. In Section~\ref{sec:Applications}, we propose a set of \gls{soa} wearable applications along with a systematic characterization of their key features. In Sections~\ref{sec:ExpSetup} and~\ref{sec:ExpRes}, we describe the setup of our experiments and analyze the results, respectively. Finally, we summarize the key findings of this work in Section~\ref{sec:Conclusion}.

\begin{figure*}[t]
    \centering
    \includegraphics[width=0.8\textwidth]{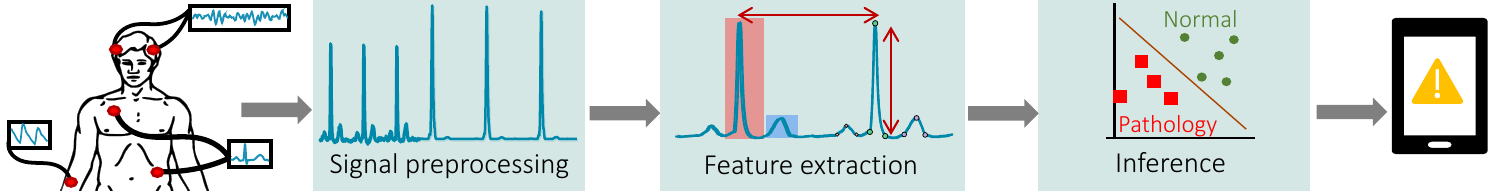}
    \caption{Typical modules of biomedical applications \cite{p:elisabetta2020}.}
    \label{fig:appModularStructure}
\end{figure*}

\begin{figure}[t]
	\centering
	\includegraphics[width=.8\linewidth]{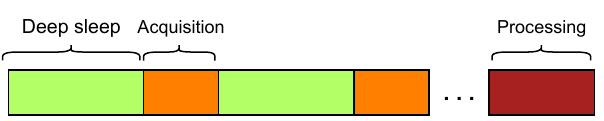}
	\caption{\gls{mcu} operating phases}
	\label{fig:mcu_operating_phases}
	\vspace{-0.45cm}
\end{figure}

\section{Related Work}\label{sec:RelatedWork}

BCIBench~\cite{BCIBench} is a benchmark suite targeting \gls{eeg}-based kernels and applications in the low-power wearables domain. Hermit~\cite{hermit} targets biomedical workloads in the \gls{iomt}. Hermit includes three kernels for monitoring common medical conditions (sleep apnea, heart-rate variability, and blood pressure), kernels for offline diagnosis using image processing, and encryption and compression algorithms. Finally, ImpBench~\cite{impbench} targets devices in the implantable biomedical domain. ImpBench features two synthetic benchmark applications for motion detection and drug delivery system simulation, and six lightweight kernels for data compression, encryption, and integrity. 

BiomedBench differs from existing biomedical benchmark suites in two ways. First, BiomedBench integrates the idle and acquisition phases in complete end-to-end applications and highlights the impact of these phases in the application and platform design. Second, BiomedBench includes a wider variety of processing kernels stemming from a larger set of input signals, like \gls{ecg}, \gls{semg}, and \gls{ppg}, thus covering a wider spectrum of workloads, as shown in Section~\ref{sec:applications:apps}.

\section{Applications}\label{sec:Applications}

Fig.~\ref{fig:appModularStructure} shows a typical biosignal monitoring application pipeline. Sensors capture the biosignal and send it to the processing device for analysis. Typically, the processing step consists of signal preprocessing (i.e., filtering), feature extraction (i.e., time or frequency characteristics), and inference (i.e., \gls{ml} model). However, applications can exhibit a wide range of workloads and computational requirements. For example, feature extraction can be implemented explicitly (i.e., features designed manually) or implicitly (e.g., \gls{cnn}). Similarly, the inference step can use a lightweight machine learning method, such as a random forest or a computationally intensive \gls{dnn}.

From an implementation point of view, the \gls{mcu} interchanges among idle, acquisition, and processing, as presented in Fig.~\ref{fig:mcu_operating_phases}. Assuming an external \gls{adc} with a buffer, \gls{mcu} collects the data through a communication protocol such as \gls{spi} upon buffer filling. The acquisition is typically served by the \gls{dma}. The \gls{mcu} is in low-power modes during idle and acquisition (i.e., deep sleep for idle and light sleep for acquisition). The duration of the low-power modes can vary significantly between applications and can dominate the system's energy consumption. Considering the variety of workloads, input bandwidths, and idle-to-active ratios present in low-power wearable applications, a benchmark suite that covers a wide range of applications is an essential tool for hardware evaluation in the domain.

\subsection{Metrics for application characterization}\label{sec:applications:app_metrics}
We propose an application characterization by metrics. {\color{myColor}We have selected the minimum number of metrics that efficiently describe} the computing profile, impact of the idle period, acquisition intensity, and memory requirements of the applications. {\color{myColor}This information is critical for identifying the challenges posed across the three phases of a complete application cycle.} To this end, we have selected the following five metrics: \textit{main operations}, \textit{duty cycle}, \textit{input bandwidth}, \textit{dynamic data}, and \textit{static data}. 

\subsubsection {Main operations}
Identifies whether the dominant operations are branches, logical operations, \gls{fxp} or \gls{fp} computations, among others. This metric hints at the microarchitectural design {\color{myColor}and is vital to interpreting the performance variations among different architectures during processing}.

\subsubsection{Duty cycle}
Represents the ratio between CPU active cycles and total cycles. A low duty cycle means that {\color{myColor}the idle phase dominates} the energy footprint. We use the following scale: ``very low'' (less than \SI{0.1}{\percent}), ``low'' (between \SIrange{0.1}{1}{\percent}), ``medium'' (between \SIrange{1}{15}{\percent}), ``high'' (between \SIrange{15}{60}{\percent}), and ``very high'' (above \SI{60}{\percent}). Since this metric is platform-dependent, we calculate it running on an ARM Cortex-M4.

\subsubsection{Input bandwidth}
Measured in \si{\byte/\sec}, is the product of the sensor's sampling rate, the size per sample, and the number of channels used. It specifies the intensity and energy impact of the signal acquisition phase.

\subsubsection{Static data}
Measured in \si{\kibi\byte}, it quantifies the memory required for the code and read-only data, such as pre-trained parameters. Hence, it defines the amount of memory retained during idle phases{\color{myColor}, which can be critical for idle consumption.} 

\subsubsection{Dynamic data}
Measured in \si{\kibi\byte}, defines how much memory the application requires during runtime for the stack and the heap. Hence, it specifies the minimum amount of RAM {\color{myColor}needed for a deployment platform.}

\begin{table*}[tp]
\begin{center}
\resizebox{0.9\textwidth}{!}{%
\begin{tabular}{l@{\hskip0.25cm}l@{\hskip0.15cm}l@{\hskip0.15cm}r@{\hskip0.15cm}r@{\hskip0.15cm}r@{\hskip0.15cm}} 
    \toprule
    \textbf{Application} &  \textbf{Main operations} & \textbf{Duty cycle} & \multicolumn{1}{l}{\textbf{Input bandwidth} \textbf{\si{(\byte/\sec)}} } & \multicolumn{1}{l}{\textbf{Static data} \textbf{(\si{\kibi\byte})}} & \multicolumn{1}{r}{\textbf{Dynamic data} \textbf{(\si{\kibi\byte})}} \\
    \midrule
    \acrshort{hcl} & Branches (\acrshort{fxp} min/max search) & Low & 1536 & 25 & 30 \\ 
    \acrshort{seizdetsvm} & 32-bit \acrshort{fxp} multiplications/divisions & Very Low & 128 & 40 & 40 \\
    \acrshort{seizdetcnn} & 16-bit \acrshort{fxp} \acrshort{mac} & High & 11776 & 350 & 120 \\
    \acrshort{cwm} & 32-bit \acrshort{fxp} multiplications & Medium & 4096 & 90 & 50 \\
    \acrshort{gcl} &  32-bit \acrshort{fp} \acrshort{mac} & Very High & 192000 & 50 & 110 \\
    \acrshort{coughdet} & 32-bit \acrshort{fp} multiplications & Very High & 64400 & 568 & 160 \\
    \acrshort{ecl} & Branches (\acrshort{fp} sorting) & Low & 822 & 16 & 4 \\
    \acrshort{bpfree} & 32-bit \acrshort{fp} \acrshort{mac} & - & - & 1300 & 2600 \\
    \bottomrule
\end{tabular}
}
\end{center}
\caption{Benchmark applications - A characterization by metrics}
\label{App_Metrics_Table}
\end{table*}

\subsection{\gls{soa} wearable applications}\label{sec:applications:apps}
We have selected eight biomedical wearable applications that offer representative workloads and varied profiles for the processing, idle, and acquisition phases. The applications are complementary and enable the evaluation of different architectural parts (e.g., sleep mode, digital signal processing). BiomedBench will be launched with eight applications but is open to future additions that present new challenges in any of the three phases.

All applications are coded {\color{myColor}and optimized} in C or C++ {\color{myColor}to ensure effortless portability across all platforms. Considering that modern C/C++ toolchains are capable of applying heavy optimizations and fully exploiting the underlying microarchitecture (i.e., \acrshort{dsp} \acrshort{isa} extensions, SIMD instructions), we consider this approach practical, consistent, and fair for comparison of \gls{soa} platforms.} 

Table~\ref{App_Metrics_Table} summarizes the main metrics of the applications, illustrating the broad spectrum of computational workloads, active-to-idle ratios, and acquisition and memory requirements covered by the benchmark suite. This variety of requirements is vital to a complete evaluation of low-power platforms, as illustrated in Section~\ref{sec:ExpRes}. The benchmarks are explained below.

\setcounter{subsubsection}{0}

\subsubsection{\Gls{hcl}} Detects abnormal heartbeat patterns in real time for common heart diseases using the \gls{ecg} signal~\cite{rpclassifier_paper}. The input signal is sampled by three different \gls{ecg} leads at \SI{256}{\hertz} with 16-bit accuracy for \SI{15}{\second}. The input signal is processed through \gls{mf}, and the \gls{rms} combines the three signal sources before enhancing the signal through \gls{relen}. In feature extraction, the \gls{reward} algorithm detects the R peaks before delineating the other fiducial points of \gls{ecg}. Finally, a neuro-fuzzy classifier using \gls{rp} of the fiducial points classifies the heartbeats as abnormal or not.
 
The application uses 16-bit fixed-point arithmetic. \gls{mf} is the dominant kernel, accounting for more than \SI{80}{\percent} of the execution time. The \gls{mf} implementation involves a queue to perform dilation and erosion, translating into data movements and min/max search. We also include the multicore version of the application~\cite{p:elisabetta2020}, having improved the parallelization strategy for the delineation and classification phases with dynamic task partition instead of static.
 
\subsubsection{\Gls{seizdetsvm}}
 Works on \gls{ecg} input and recognizes epileptic episodes in real time~\cite{Farnaz_SVM}. The \gls{ecg} signal is sampled from a single lead at \SI{64}{\hertz} with 16-bit accuracy for \SI{60}{\second}. The preprocessing phase consists of a simple \gls{ma} subtraction. For feature extraction, the \gls{rri} and \gls{edr} time series are calculated from the \gls{ecg}. From \gls{rri}, \gls{hrv} features and Lorenz plot features are extracted. From \gls{edr}, the linear predictive coefficients and the power spectral density of different frequency bands are calculated. For the \gls{ffe} of \gls{rri} and \gls{hrv}, the \gls{plomb} algorithm, which involves a \gls{fft}, is used. For inference, a \gls{svm} uses all the extracted features to classify the patient's state.

\gls{plomb} is the dominant kernel, accounting for more than \SI{75}{\percent} of the execution time. Since the implementation is in 32-bit \gls{fxp} arithmetic, the main operations are 32-bit integer multiplications with a 64-bit intermediate result followed by a shift. This application originally contained a self-aware mechanism to determine the number of features and the complexity of the \gls{svm}. For this benchmark, we only use the full pipeline to avoid variability among executions and test the most complete version. Finally, we designed a parallel version of this application since it features a high degree of parallelism.

\subsubsection{\Gls{seizdetcnn}}
Based on \gls{eeg} data, detects epileptic seizure episodes in real time~\cite{Gmez2020AutomaticSD}. The signal is sampled from 23 leads at \SI{256}{\hertz} with a 16-bit accuracy for \SI{4}{\second}. This application does not feature any signal preprocessing or feature extraction kernels. Instead, the input is sent directly to the input layer of a \gls{fcn}. The proposed \gls{fcn} architecture has three 1D convolutional layers, each including batch normalization, pooling, and ReLU layers, and two fully connected layers. Most computations are 16-bit \gls{fxp} \gls{mac} operations due to convolution, as \SI{90}{\percent} of the execution is spent in the convolutional layers. {\color{myColor} We implemented the \gls{fcn} in C from scratch for single-core and multicore platforms.}

\subsubsection{\Gls{cwm}}
Is designed for real-time monitoring of the cognitive workload state of a subject~\cite{j:zanetti2022} and is based on \gls{eeg} input. The \gls{eeg} signal is sampled by four leads at \SI{256}{\hertz} with 32-bit accuracy. The input signal is processed in 14 batches of \SI{4}{\second} for a total of \SI{56}{\second}. Preprocessing and feature extraction are executed 14$\times$ per channel before the classification phase is executed. Preprocessing involves \gls{blr} and a \gls{bpf} through \gls{iir} filters. Feature extraction contains time-domain features (i.e., skewness/kurtosis, Hjorth activity), frequency-domain features (i.e., power spectral density), and entropy features. A \gls{rf} uses these features to classify the stress condition of the subject. 

The extraction of frequency features, which contains the \gls{fft}, is the most demanding computational kernel, accounting for more than \SI{80} {\percent} of the total computation time. The main operations are 32-bit integer multiplications with a 64-bit intermediate result followed by a shift since we transformed the original application into a \gls{fxp} implementation with a negligible accuracy drop.

\subsubsection{\Gls{gcl}}
Aims to classify hand gestures by inspecting signals captured by \gls{semg} of the forearm~\cite{BSS}. The idea is to extract the \gls{muapt} and identify the motor neuron activity patterns to classify the hand gesture. The signal is sampled from 16 channels at \SI{4}{\kilo\hertz} with 24-bit accuracy for only \SI{0.2}{\second}. This application has no signal preprocessing. The authors apply a \gls{bss} method to the input signal, namely \gls{ica}, and classify the gesture using a \gls{svm} or a \gls{mlp}. We use the \gls{mlp} for the inference stage to boost the variability of the kernels under test. 

\gls{gcl} is implemented in 32-bit \gls{fp} arithmetic, and the dominant workload is the \gls{ica} which features matrix multiplications. Hence, the main operations are 32-bit \gls{fp} \glspl{mac}. We have included the original parallel implementation of this application and converted it to run on a single core to make it available for single-core platforms. 

\subsubsection{\Gls{coughdet}}
Is a novel application~\cite{orlandic_multimodal_2023} using non-invasive chest-worn biosensors to count the number of cough episodes people experience per day, thus providing a quantifiable means of evaluating the efficacy of chronic cough treatment. The device records audio data, sampled at \SI{16}{\kilo\hertz} with 32-bit precision, as well as 3-axis accelerometer and 3-axis gyroscope signals from an \gls{imu}, each sampled at 100 Hz with 16-bit precision. Biosignals are processed every \SI{0.3}{\second}. 


Feature extraction includes computations in the time and frequency domain. Time-domain computations include the extraction of statistical values (such as zero crossing rate, root-means-squared, and kurtosis) of the \gls{imu} signals. 
An \gls{fft} is used to extract spectral statistics (including standard deviation and dominant frequency), power spectral density, and \gls{mfcc} of the audio signal. Features extracted from audio and \gls{imu} signals are forwarded to an \gls{rf} classifier that computes the probability of a cough event.

The \gls{mfcc} constitutes the most intensive kernel that requires the iterative computation of \gls{fft} and transcendental functions (i.e., the logarithm in the \gls{dct}). The application is implemented in 32-bit \gls{fp} arithmetic, and the main operations include \gls{fp} multiplications. {\color{myColor}We coded this application from scratch.}

\subsubsection{\Gls{ecl}}
Classifies the fear status of patients to prevent gender-based violence \cite{Jose_bindi} based on three physiological signals, namely \gls{gsr}, \gls{ppg}, and \gls{st}. \gls{ppg} is sampled at \SI{200}{\hertz} with 32-bit precision, \gls{gsr} is sampled at \SI{5}{\hertz} with 32-bit precision, and \gls{st} is sampled at \SI{1}{\hertz} with 16-bit precision. The acquisition window lasts \SI{10}{\second} and is divided into 10 batches of partial inference before the final classification is performed based on the 10 partial classifications.
 
\gls{ecl} has no signal preprocessing step. Feature extraction includes the \gls{avg} of the three input signals over \SI{1}{\second} before forwarding them to a \gls{knn} classifier. The classifier computes the distances of the new 3D tuple from the training points that have already been labeled as fear or no fear. Using $n$ training points, we select the $\sqrt{n}$ closest training points by running $\sqrt{n}$ steps of selection sort before classifying the new tuple based on the percentage of neighboring fear-labeled points. We use 685 training points---a tradeoff between accuracy and complexity~\cite{Jose_bindi}.

\gls{ecl} uses both 16-bit and 32-bit \gls{fp} arithmetic, as it uses different representations for the three different signals. The dominant kernel is the \gls{knn} inference, which includes the 32-bit \gls{fp} calculation and sorting of the Euclidean distances in 3D. Sorting includes multiple minimum search iterations over the array of distances.

\subsubsection{\Gls{bpfree}}
Is {\color{myColor}a neural network training scheme for resource-constrained devices, that is ideal for on-device training scenarios where personalized samples remain private and can increase the model's robustness \cite{biobpfree}}. The main notion of \gls{bpfree} is to perform per-layer training by maximizing the distance between the intermediate outputs of different classes and minimizing the distance between the intermediate outputs of the same class. \gls{bpfree} avoids the prohibitive memory cost of backpropagation, thus opening possibilities for on-device training in low-power devices.

We used \gls{bpfree} to fine-tune the \gls{dnn} presented in~\cite{Gmez2020AutomaticSD} for seizure detection. The model was originally trained on the server using a leave-one-out-patient on the CHB-MIT database. Later, we retrain the model on the device with \gls{bpfree} by exploiting the personalized samples acquired from the patient under test. The on-device training yields a significant improvement in the F1 score up to 25\% thanks to the personalized samples available on the device while ensuring data privacy.

The implementation of \gls{bpfree} is based on the computation of the gradients of the loss function with respect to the trainable parameters. We define a custom loss function per layer~\cite{biobpfree} and then compute the gradients using the chain rule to account for the intermediate layers (i.e., ReLU, batch normalization, max pooling). The main operations are 32-bit \gls{fp} \glspl{mac} because of the convolution in the forward passes and the vector-matrix multiplications involved in the chain rule of the gradient computation. There is no acquisition phase. We assume that four pre-recorded input samples are already stored in the FLASH and expect the retraining to occur during the device charging phase. One epoch is executed for benchmark purposes. 

{\color{myColor}We have published all the code in a public GitHub repository,\footnote{https://github.com/esl-epfl/biomedbench} which contains detailed information about each implementation. We have also developed a website\footnote{https://biomedbench.epfl.ch/} to enhance the readability and communication of results across the community.}
\section{Experimental Setup}\label{sec:ExpSetup}
In this section, we show the deployment and evaluation process of BiomedBench on a set of representative \gls{soa} low-power boards. 

\begin{table*}[t]
\centering
\resizebox{.8\textwidth}{!}{%
\begin{tabular}{llllrrrr}
\hline
\textbf{\textbf{Board}} &
\textbf{\textbf{Manufacturer}} &
  \textbf{\textbf{MCU}} &
  \textbf{\textbf{Cores}} &
  \textbf{\textbf{FPU}} &
  \textbf{\textbf{RAM}} \textbf{(\si{\kibi\byte})} &
  \textbf{\textbf{FLASH}} \textbf{(\si{\mega\byte})}  \\
 \hline
Raspberry Pi Pico &
Raspberry &
  RP2040 &
  2x ARM Cortex-M0+ &
  No &
  264 &
  2 (off-chip)\\
Nucleo L4R5 &
STMicroelectronics &
  STM32L4R5ZI &
  1x ARM Cortex-M4 &
  Yes &
  640 &
  2 (on-chip)\\
Ambiq Apollo 3 &
Ambiq &
  Apollo 3 Blue &
  1x ARM Cortex-M4 &
  Yes &
  384 &
  1 (on-chip)\\
Gapuino &
GreenWaves &
  GAP8 &
  1x CV32E40P (FC) &
  No &
  512 &
  2 (off-chip) \\
 &
 Technologies &
   &
  8x CV32E40P (Cluster) &
  Yes &
  64 & \\
GAP9EVK &
GreenWaves &
  GAP9 &
  1x CV32E40P (FC) &
  Yes &
  1564 &
  2 (off-chip) \\
 &
  Technologies &
   &
  9x CV32E40P (Cluster) &
  Yes &
  128 &\\ \hline
\end{tabular}%
}
\caption{Selected boards - Summary of basic features}
\label{Boards_table}
\end{table*}

\subsection {Considered low-power boards}\label{sec:ExpSetup:Boards}
We target low-power platforms with low-end \glspl{mcu}, with clock frequencies in the range of MHz, and RAM in the range of some hundreds of \SI{}{\kibi\byte}, as the application characterization in Table~\ref{App_Metrics_Table} suggests. Typically, such platforms feature simple processor architectures with in-order execution, no instruction-level parallelism, simple memory hierarchies, deep-sleep modes for long idle periods, and \gls{spi} support for signal acquisition. Such platforms often meet the application requirements in the domain~\cite{p:elisabetta2020}, \cite{Farnaz_SVM}, \cite{j:zanetti2022}, \cite{BSS}. Finally, we explicitly target variability in processor architectures (i.e., ARM, RISC-V).
 
We have selected five popular commercial low-power boards featuring five different \glspl{mcu} and four different processors for our experiments. The selected boards are: Nucleo-L4R5ZI\footnote{https://www.st.com/en/evaluation-tools/nucleo-l4r5zi.html} from ST Microelectronics, Ambiq Apollo~3 Blue\footnote{https://ambiq.com/apollo3-blue/}, Raspberry Pi Pico\footnote{ https://www.raspberrypi.com/products/raspberry-pi-pico/}, Gapuino v1.1\footnote{https://greenwaves-technologies.com/product/gapuino/} and GAP9EVK\footnote{https://greenwaves-technologies.com/gap9-store/} from GreenWaves Technologies. We summarize the architecture and storage specifications of each \gls{mcu} in Table~\ref{Boards_table}.

\subsection {Sensor emulation and sleep modes}\label{sec:ExpSetup:Deployment}
For signal acquisition, we emulate the sensor and \gls{adc} functionality using an external board that artificially produces data. We assume an external \gls{adc} with 768 bytes of RAM buffer\footnote{AD4130-8, https://www.analog.com/en/products/ad4130-8.html} since, typically, \glspl{mcu} do not feature embedded \glspl{adc} or feature \glspl{adc} with insufficient bit precision. We perform per-batch acquisition by transferring data to the \gls{mcu} when the buffer is full. We employ an \gls{spi} acquisition scheme using the \gls{dma} while the core is in sleep mode. Moreover, we assume that the sensors have no processing ability and that all the computations take part in the \gls{mcu}. 

During the idle period, we set the \gls{mcu} to deep-sleep mode with RAM retention to preserve the data needed for the next processing cycle. The \glspl{mcu} support a wake-up interrupt mechanism to switch to active mode when the data are ready. For the processing phase, we select the lowest operating voltage that allows the processing frequency to meet the real-time constraints of each application. For the selected voltage, we configure the highest available frequency for maximum energy efficiency as validated experimentally.

\subsection {Energy measurements}\label{sec:ExpSetup:Energy}
We use the evaluation boards provided by the manufacturers to measure the energy consumption of each \gls{mcu} executing BiomedBench applications. We do not consider the energy of the sensor and \gls{adc} in our experiments, as it is common for all platforms. We have measured the energy consumption of all boards at the power supply entry point of the \gls{ic}\footnote{We measure the total board energy for Raspberry Pi Pico with all peripherals disabled --- no test points for the \gls{mcu} provided} as we target a fair comparison across all platforms. However, we highlight that the reported energy numbers include the energy drawn by the input-to-core step-down voltage converter embodied in the integrated circuit. Future platform energy measurements must comply with this procedure for the results to be considered valid.

We have selected the Otii Arc provided by Qoitech, which samples at \SI{4}{\kilo\hertz}, to obtain an energy profile over time and extract the energy and execution time per phase. However, due to the limited $\pm$\SI{10}{\micro\ampere} precision of the Otii device, we used the Fluke 8846A~multimeter to measure the average current of the Nucleo and Apollo boards in deep-sleep mode. This device can achieve a precision of \SI{0.03}{\micro\ampere}.

\subsection{Portability and software support}\label{sec:ExpSetUp:portability}
We use the boards' toolchain and \gls{sdk} to compile {\color{myColor}with -O3} and load the C/C++ program on each board. {\color{myColor}With this approach, we ensure that \gls{dsp} extensions are exploited when present}. For the runtime, we use the portable FreeRTOS API, which all boards support, for dynamic memory management. Finally, we utilize the \gls{hal} of each board's \gls{sdk} to program the \gls{spi} peripheral communication and to configure the \gls{pmu} for the sleep modes. 
\section{Experimental Results}\label{sec:ExpRes}

\begin{table*}[t]
\begin{center}

\resizebox{1.9\columnwidth}{!}{%
\begin{tabular}{cc|rrrrrr|rrrrrr}
    \toprule
    \multirow{2}{*}{\textbf{MCU}} &
    \multirow{2}{*}{\textbf{Processor}} &
    \multirow{2}{*}{\textbf{Application}} &
    \multirow{2}{*}{\makecell{\textbf{Cycles} \\ (\si{\textbf{\mega}})}} &
    \multicolumn{4}{c|}{\textbf{Energy (\si{\textbf{\milli\joule}})}} &
    \multirow{2}{*}{\textbf{Application}} &
    \multirow{2}{*}{\makecell{\textbf{Cycles} \\ (\si{\textbf{\mega}})}} &
    \multicolumn{4}{c}{\textbf{Energy (\si{\textbf{\milli\joule}})}} 
    \rule{0pt}{12pt} 
    \\
    \cline{5-8} \cline{11-14}
    \rule{0pt}{12pt} 
    & & & &
     \textbf{Idle} & \textbf{Acq.} & \textbf{Proc.} & \textbf{Total}
    & & &
    \textbf{Idle} & \textbf{Acq.} & \textbf{Proc.} & \textbf{Total}
     \\ 
    \midrule
    RP2040 & Arm Cortex-M0+
    & \multirow{5}{*}{\gls{hcl}} & 11.6 & 29.647 & 3.519 & 6.532 & 39.698
    & \multirow{5}{*}{\gls{coughdet}} & 149.7 & 0 & 2.972 & 87.543 & 90.515 \\
    STM32L4R5ZI & Arm Cortex-M4 &  
    & 7.4 & 0.118 & 0.002 & 2.604 & 2.724
    &
    & 9.9 & 0.002 & 0.001 & 6.649 & 6.652 \\
    Apollo 3 Blue & Arm Cortex-M4 &  
    & 7.4 & 0.073 & 0.061 & 0.226 & 0.360
    &
    & 9.9 & 0.001 & 0.198 & 0.444 & 0.642 \\
    GAP8 & CV32E40P GAP8 & 
    & 5.1 & 9.386 & 0.042 & 0.416 & 9.844
    &
    & - & - & - & - & - \\
    GAP9 & CV32E40P GAP9 & 
    & 5.1 & 9.833 &	0.154 &	0.411 &	10.398
    &
    & 9.1 & 0.081 &	0.046 &	0.352 &	0.479 \\
    \midrule
    
    RP2040 & Arm Cortex-M0+
    & \multirow{5}{*}{\gls{seizdetsvm}} & 4.3 & 118.805 & 2.402 & 2.420 & 123.627
    & \multirow{5}{*}{\gls{gcl}} & 571.6 & 0 & 8.008 & 347.104	& 355.112 \\
    STM32L4R5ZI & Arm Cortex-M4 &  
    & 2.3 & 0.476 & 0.001 & 1.313 & 1.790
    &
    & 23.0 & 0 & 0.004 & 13.425 & 13.429 \\
    Apollo 3 Blue & Arm Cortex-M4 &  
    & 2.3 & 0.294 & 0.042 & 0.137 & 0.473
    &  
    & 23.0 & 0 & 0.525 & 2.500 & 3.025 \\
    GAP8 & CV32E40P GAP8 &  
    & 2.8 & 37.724 & 0.029 & 0.353 & 38.106
    &  
    & 635.8 & 0 & 0.096 & 220.933 & 221.029\\
    GAP9 & CV32E40P GAP9 &  
    & 2.5 & 39.50 & 0.037 & 0.090 & 39.627
    &  
    & 20.2 & 0.027 & 0.124 & 0.604 & 0.755\\
    \midrule
    
    RP2040 & Arm Cortex-M0+ 
    & \multirow{5}{*}{\gls{seizdetcnn}} & 283.0 & 3.514 & 7.528 & 167.87&178.912
    & \multirow{5}{*}{\gls{ecl}} & 15.3 & 19.651 & 1.259 & 8.760 & 29.670 \\
    STM32L4R5ZI & Arm Cortex-M4 &  
    & 240.0 & 0.015	& 0.004 & 112.049 & 112.068
    &  
    & 2.5 & 0.002 & 0.001 & 1.462 & 1.465\\
    Apollo 3 Blue & Arm Cortex-M4 &  
    & 240.0 & 0.010 & 0.494 & 18.262 & 18.766

    &  
    & 2.5 & 0.061 & 0.083 & 0.110 & 0.254\\
    GAP8 & CV32E40P GAP8 &  
    & 160.0 & 0.464 & 0.090 & 31.987 & 32.541
    &  
    & 14.3 & 6.224 & 0.015 & 1.052 & 7.291\\
    GAP9 & CV32E40P GAP9 &  
    & 160.0 & 2.234 & 0.117 & 5.101 & 7.452
    &  
    & 1.6 & 6.572 & 0.019 & 0.061 & 6.652 \\
    \midrule
    
    RP2040 & Arm Cortex-M0+ 
    & \multirow{5}{*}{\gls{cwm}} & 346.0 & 104.902&	35.876&	195.910& 336.688
    & \multirow{5}{*}{\gls{bpfree}} & 16758.0 & - & - & 9374.500 & 9374.500\\
    STM32L4R5ZI & Arm Cortex-M4 &  
    &138.0 & 0.432 & 0.017 & 70.629 & 71.078
    &  
    &662.0 & - & - & 432.227 & 432.227\\
    Apollo 3 Blue & Arm Cortex-M4 &  
    &138.0 & 0.325 & 0.620 & 3.887 & 4.832
    &  
    &662.0 & - & - & 32.222 & 32.222\\
    GAP8 & CV32E40P GAP8 &  
    &165.0 & 33.930 & 0.431 & 16.008 & 50.369
    &  
    &18450.0 & - & - & 1453.368 & 1453.368\\
    GAP9 & CV32E40P GAP9 &  
    &92.0 & 36.303 & 0.557 & 3.711 & 40.571
    &  
    &633.0 & - & - & 24.970 & 24.970 \\
    \bottomrule
\end{tabular}%
}
\end{center}
\caption{Energy breakdown and processing cycles per application.}
\label{results_table}
\end{table*}

In this section, we evaluate \gls{soa} platforms running BiomedBench in terms of energy efficiency and processing capability. We showcase that BiomedBench stresses different architectural aspects of the platforms, hence making it an effective hardware evaluation tool. Table~\ref{results_table} reports the processing cycles and energy per application and board.

\subsection{Processing cycles}
The amount of processing cycles required to execute the computational phase of the applications is a critical metric for evaluating wearable devices. Fewer processing cycles translate to shorter active phases for the \gls{mcu} and energy efficiency. Analyzing in depth the processing performance discrepancies of the \gls{soa} platforms is the key to comprehending the exact microarchitectural challenges in the domain and, hence, facilitating domain-specific hardware design.

We summarize our observations stemming from Table~\ref{results_table} in four key points. First, CV32E40P GAP9 consistently scores the highest in all applications. Second, the relative performance of CV32E40P GAP8 and Arm Cortex-M4 varies significantly depending on the application type. In some applications, Arm Cortex-M4 outperforms CV32E40P GAP8 and matches CV32E40P GAP9, while in other applications, CV32E40P GAP8 outperforms Arm Cortex-M4 and matches CV32E40P GAP9. Third, Arm Cortex-M0+ cannot handle computations as efficiently as the other processors. Finally, Arm Cortex-M0+ and CV32E40P GAP8 lack a \gls{fpu} and suffer in \gls{fp} applications.

\subsection{Energy}
Understanding why some platforms are more energy-efficient than others and how the energy profile changes among different applications and their phases is vital to boosting low-power platform design. Table~\ref{results_table} reports the total energy and the energy per phase for each application and platform. Interestingly, the impact of the phases on the total energy footprint fluctuates with the application.

\subsubsection{Idle}
Low-duty-cycle applications highlight the importance of a well-designed deep-sleep mode. \gls{seizdetsvm}, featuring the lowest duty cycle of all applications, illustrates that STM32L4R5ZI and Apollo 3 Blue have excellent deep-sleep modes and dominate their rivals in total energy. Similar observations apply to \gls{hcl} and \gls{ecl}. In contrast, idle energy is much less impactful in high-duty-cycle applications such as \gls{seizdetcnn}, \gls{gcl}, and \gls{coughdet}.

\subsubsection{Acquisition}
Applications with a high input bandwidth highlight the need for an energy-efficient acquisition mode. \gls{coughdet} and \gls{gcl}, featuring the highest input bandwidth, stress the importance of an energy-efficient acquisition mode (i.e., a sleep mode that allows \gls{dma} operation). Apart from Apollo 3, all platforms employ the \gls{dma} and spend negligible energy during acquisition in \gls{coughdet} and \gls{gcl}. The acquisition energy is very low for all other applications.

\subsubsection{Processing}
High-duty-cycle applications, like \gls{seizdetcnn}, \gls{coughdet}, and \gls{gcl}, necessitate an energy-efficient processing mode. Each application features a different computational workload that impacts the duration of the processing phase per platform. The platforms' relative processing efficiency fluctuates per application, depending on their processor's performance across different workloads. 

In general, Apollo 3 and GAP9 have the lowest processing energy. However, GAP9 manages to outperform Apollo 3 in applications that it can complete in significantly fewer cycles (i.e., \gls{seizdetcnn}, \gls{ecl}). GAP8 spends more processing energy than GAP9 even when it matches GAP9's processing cycles (i.e., \gls{hcl}, \gls{seizdetcnn}). Despite featuring the same processor, STM32L4R5ZI consumes approximately an order of magnitude more processing energy than Apollo 3 Blue in all applications. RP2040 spends, on average, two orders of magnitude more processing energy than the most efficient \gls{mcu}. 

\subsubsection{Total}
{\color{myColor}The experimental results show that} no single platform is the most energy-efficient for every benchmark. GAP9 is the most energy-efficient in computationally intensive, high-duty-cycle applications but features an uncompetitive deep-sleep mode. Apollo~3 and STM have the best deep-sleep modes and perform the best in low-duty-cycle applications. {\color{myColor}Therefore, total energy consumption varies significantly for each platform according to the characteristics of each application.} For instance, STM is \SI{22}{\times} more energy-efficient than GAP9 in the \gls{seizdetsvm}{\color{myColor}, which is a low duty cycle application, and thus the idle mode efficiency of STM is more relevant}, but has \SI{23.5}{\times} more energy consumption in \gls{seizdetcnn}, {\color{myColor}which is a high duty cycle application that benefits the processing efficiency of GAP9. The selected applications feature diverse requirements and pose different challenges for low-power platforms, making BiomedBench a representative benchmark for evaluating architectural designs in the TinyML wearable domain.}

\section{Conclusion}\label{sec:Conclusion}
In this paper, we have introduced BiomedBench, a new {\color{myColor}biomedical} benchmark suite aiming to systematize {\color{myColor}hardware evaluation in the TinyML wearable domain.} BiomedBench features end-to-end applications with diverse computational pipelines, active-to-idle ratios, and acquisition profiles. BiomedBench, boosted by a systematic application characterization, unveils the key hardware challenges of deploying modern biomedical applications during idle, acquisition, and processing phases on wearable platforms. By evaluating BiomedBench’s impact on energy and performance for five \gls{soa} low-power platforms, we have shown that no single \gls{mcu} can efficiently handle the varying challenges of different benchmarks. To this end, BiomedBench will be released as an open-source suite to systematize platform evaluation, accelerate hardware design, and enable further advances in bioengineering systems and TinyML application design.
\section*{Acknowledgements}

This work was supported in part by the Swiss State Secretariat for Education, Research, and Innovation (SERI) through the SwissChips research project, by the Wyss Center for Bio and Neuro Engineering: Lighthouse Noninvasive Neuromodulation of Subcortical Structures, by ACCESS – AI Chip Center for Emerging Smart Systems, sponsored by InnoHK funding, Hong Kong SAR, and by the Swiss NSF Sinergia, grant no. 193813: ”PEDESITE-Personalized Detection of Epileptic Seizure in the Internet-of-Things Era.


%




\ifCLASSOPTIONcaptionsoff
  \newpage
\fi

\vspace{-2\baselineskip}
\begin{IEEEbiographynophoto}{Dimitrios Samakovlis}
is a Ph.D. student in the Embedded Systems Laboratory (ESL) at EPFL (Lausanne, Switzerland). He received his Master's in Electrical and Computer Engineering (2021) from the University of Thessaly, Volos, Greece. His research focuses on characterizing and deploying biomedical wearable applications in low-power platforms, on-device training on resource-constrained devices, and strategies for efficient synergy between Edge devices and servers. 
\end{IEEEbiographynophoto}

\vspace{-2\baselineskip}

\begin{IEEEbiographynophoto}{Stefano Albini}
is a Ph.D. student in the Embedded Systems Laboratory (ESL) at EPFL (Lausanne, Switzerland). He received his Master's degree in Computer Engineering from the University of Pavia, Italy, in 2022.
His research focuses on deploying biomedical applications on wearable nodes and optimizing AI techniques for constrained edge devices.
\end{IEEEbiographynophoto}

\vspace{-2\baselineskip}

\begin{IEEEbiographynophoto}{Rubén Rodríguez Álvarez} is pursuing his PhD in the Embedded Systems Laboratory at EPFL in Switzerland. His research interest is in SW-HW exploration and co-design using domain-specific accelerators for edge and cloud computing. He completed his master's in industrial and electronics engineering at the Technical University of Madrid in 2021 after receiving his bachelor's in industrial engineering from Carlos III University in Madrid in 2019. He actively contributes to the development of an open-source hardware platform called X-HEEP and a CGRA accelerator for the edge.
\end{IEEEbiographynophoto}

\vspace{-2\baselineskip}

\begin{IEEEbiographynophoto}{Denisa-Andreea Constantinescu}
is a PostDoc at the Embedded Systems Laboratory at EPFL in Switzerland, where she is leading research on sustainable computing technologies for urban digital twins and the development of energy-aware hardware accelerators for the SKA Observatory. She obtained her Ph.D. in Mechatronics in 2022 and her Master's in Computer Engineering in 2017 from UMA, Spain. She completed two research stays: at the EPFL Embedded Systems Laboratory (Lausanne, Switzerland) in 2022 and at the Northeastern University Computer Architecture Research Laboratory (Boston, USA) in 2018. In recognition of her work, she received the Intel oneAPI Innovator award in 2020 for her research on ``Efficiency and Productivity for Decision-making on Mobile SoCs" and the SCIE-ZONTA Award from the Scientific Society of Informatics in Spain in 2021. 
\end{IEEEbiographynophoto}

\vspace{-2\baselineskip}

\begin{IEEEbiographynophoto}{Pasquale Davide Schiavone} is a PostDoc at the EPFL and Director of Engineering of the OpenHW Group. He obtained the Ph.D. title at the Integrated Systems Laboratory of ETH Zurich in the Digital Systems group in 2020 and the BSc. and MSc. from ``Politecnico di Torino" in computer engineering in 2013 and 2016, respectively. His main activities are the RISC-V CPU design and low-power energy-efficient computer architectures for smart embedded systems and edge-computing devices.
\end{IEEEbiographynophoto}

\vspace{-2\baselineskip}

\begin{IEEEbiographynophoto}{Miguel~Pe\'{o}n-Quir\'{o}s}
received a Ph.D. in Computer Architecture from UCM, Spain, in 2015. He collaborated as a Marie Curie scholar with IMEC (Leuven, Belgium) and as a postdoctoral researcher with IMDEA Networks (Madrid, Spain) and the Embedded Systems Laboratory (ESL) at EPFL (Lausanne, Switzerland). He has participated in several H2020, SNSF, and industrial projects and is currently part of EcoCloud, EPFL. His research focuses on energy-efficient computing.
\end{IEEEbiographynophoto}

\vspace{-2\baselineskip}

\begin{IEEEbiographynophoto}{David~Atienza}
(M'05-SM'13-F'16) is a Professor of electrical and computer engineering, Heads the Embedded Systems Laboratory (ESL), and is the Scientific Director of the EcoCloud Center for Sustainable Computing at the École Polytechnique Fédérale de Lausanne (EPFL), Switzerland. He received his Ph.D. in computer science and engineering from UCM, Spain, and IMEC, Belgium, in 2005. His research interests include system-level design methodologies for high-performance multi-processor system-on-chip (MPSoC) and low-power Internet-of-Things (IoT) systems, including new 2-D/3-D thermal-aware design for MPSoCs and many-core servers, and edge AI architectures for wearable systems and smart consumer devices. He is a co-author of more than 400 papers in peer-reviewed international journals and conferences, one book, and 14 patents in these fields. Dr. Atienza is an IEEE Fellow and an ACM Fellow.
\end{IEEEbiographynophoto}




\end{document}